\documentclass[mlabstract]{jmlr}

\usepackage{placeins}
\usepackage{longtable}% for long tables

\usepackage{booktabs}
\usepackage[load-configurations=version-1]{siunitx} % newer version
\usepackage{enumitem}
\usepackage{cleveref}

\newcommand{\R}{\mathbf R}
\newcommand{\dd}{\mathrm d}
\newcommand{\pos}{\mathrm p}
\newcommand{\mom}{\mathrm m}
\DeclareMathOperator{\id}{id}

\theorembodyfont{\upshape}
\theoremheaderfont{\scshape}
\theorempostheader{:}
\theoremsep{\newline}

\jmlrvolume{}
\firstpageno{1}
\editors{}

\jmlryear{2024}
\jmlrworkshop{Symmetry and Geometry in Neural Representations}

\title[Hamiltonian Matching with SympFlow]{Hamiltonian Matching for Symplectic Neural Integrators}

\author{\Name{Priscilla Canizares} \Email{pc464@cam.ac.uk}\\
\Name{Davide Murari} \Email{dm2011@cam.ac.uk}\\
\Name{Carola-Bibiane Sch\"onlieb} \Email{cbs31@cam.ac.uk}\\
\Name{Ferdia Sherry} \Email{fs436@cam.ac.uk}\\
\Name{Zakhar Shumaylov} \Email{zs334@cam.ac.uk}\\
\addr Department of Applied Mathematics and Theoretical Physics, University of Cambridge, Wilberforce Road, Cambridge CB3 0WA, UK}
\begin{document}

\maketitle

\begin{abstract}
Hamilton's equations of motion form a fundamental framework in various branches of physics, including astronomy, quantum mechanics, particle physics, and climate science. Classical numerical solvers are typically employed to compute the time evolution of these systems. However, when the system spans multiple spatial and temporal scales numerical errors can accumulate, leading to reduced accuracy. To address the challenges of evolving such systems over long timescales, we propose SympFlow, a novel neural network-based symplectic integrator, which is the composition of a sequence of exact flow maps of parametrised time-dependent Hamiltonian functions. This architecture allows for a backward error analysis: we can identify an underlying Hamiltonian function of the architecture and use it to define a \emph{Hamiltonian matching} objective function, which we use for training. In numerical experiments, we show that SympFlow exhibits promising results, with qualitative energy conservation behaviour similar to that of time-stepping symplectic integrators.
\end{abstract}

\section{Introduction}
\label{sec:introduction}
In this work, we will study the use of neural networks to integrate Hamiltonian systems, which were first defined in the context of classical mechanics and which have since found many applications in physics \citep{arnoldMathematicalMethodsClassical1978}. More specifically, when we speak of Hamiltonian systems, we are considering ordinary differential equations (ODEs) of the following form for a state variable $x\in \R^{2d}$ and \emph{Hamiltonian} function $H:\R^{2d}\to \R$:
\begin{equation}
\frac{\dd x}{\dd t} = \mathbf J \nabla H(x),\quad\text{with}\quad \mathbf J = \begin{pmatrix}0 & \id_d \\ -\id_d & 0\end{pmatrix}.
\label{eq:hamiltonian_ode}
\end{equation}
Typically, the state variable is partitioned into a position $q\in \R^d$ and momentum $p\in\R^d$. Under standard (non-restrictive) assumptions on $H$ \citep{arnoldOrdinaryDifferentialEquations1991}, the corresponding initial value problem has a unique solution for any initial condition and initial time, which can be used to define the corresponding \emph{flow map} $\phi_H:\R\times \R^{2d} \to \R^{2d}$ by 
\begin{equation}\frac{\dd}{\dd t} \phi_{H, t}(x) = \mathbf J \nabla H(\phi_{H, t}(x))\quad\text{and}\quad \phi_{H, 0}(x)= x.\label{eq:hamiltonian_flow_map}
\end{equation}
Since the exact flow map in \cref{eq:hamiltonian_flow_map} is generally not accessible, it is necessary to make a ``satisfactory'' approximation to such an exact flow map. Depending on the context, this can have different meanings, but for Hamiltonian systems, there are important structural properties that can provide guidance: the flow map is \emph{symplectic}, meaning that the Jacobian matrix $D\phi_{H, t}(x)$ satisfies the identity $[D\phi_{H, t}(x)] ^\top\mathbf J [D\phi_{H, t}(x)] = \mathbf J,$ and the Hamiltonian $H$ is conserved, i.e.\ $H(\phi_{H, t}(x)) = H(x)$. Symplecticity also implies that volumes in phase space are preserved. When numerically approximating the flow map, it is desirable to take these structural properties into account, and doing so has led to the celebrated field of \emph{geometric numerical integration} \citep{hairerGeometricNumericalIntegration2006}.

The neural network architecture that we propose takes inspiration from the methods of geometric numerical integration to define a time-dependent symplectic neural network architecture, SympFlow, that can be used to approximate Hamiltonian flow maps. There are similarities between SympFlow and previous works on symplectic neural networks \citep{jinSympNetsIntrinsicStructurepreserving2020, burbyFastNeuralPoincar2021}, but SympFlow differs from these approaches by incorporating a time dependence which allows an underlying Hamiltonian to be identified.

In this work, we study SympFlow from the perspective of physics-informed machine learning \citep{karniadakisPhysicsinformedMachineLearning2021}: we design an unsupervised learning approach for approximating Hamiltonian flow maps. It is in principle also possible to go in the other direction, fitting the approximate flow map to trajectories, and extracting the underlying Hamiltonian to discover a physical model, a task which has previously been studied in the scientific machine learning literature \citep{bertalanLearningHamiltonianSystems2019, greydanusHamiltonian2019}.

\section{Methodology}

\subsection{The SympFlow architecture}
Fundamentally, SympFlow is defined by iterated composition of exact flow maps of time-dependent Hamiltonians, each of which depends either on position or momentum, but not both. Given an arbitrary $C^2$ function $V_{\pos}:\R\times \R^d \to \R$, we can consider the map
\begin{equation}\phi_{\pos,t}((q,p))=\begin{pmatrix}q \\ p - (\nabla_q V_{\pos}(t, q) - \nabla_q V_{\pos}(0, q))\end{pmatrix},\label{eq:pos_flow_map}\end{equation}
which is the flow map (starting from time $0$) corresponding to the Hamiltonian $    H_{\pos, t}((q, p)) = \dot V_{\pos}(t, q),$ where $\dot{V}_{\pos}$ stands for $\frac{\dd}{\dd t}V_{\pos}$ and the subscript p indicates that the Hamiltonian depends on position, but not momentum. Similarly, for a $C^2$ function $V_{\mom}:\R\times \R^d \to \R$, we can consider the map 
\begin{equation}
    \phi_{\mom,t}((q,p))=\begin{pmatrix}q + (\nabla_p V_{\mom}(t, p) - \nabla_p V_{\mom} (0, p))\\ p\end{pmatrix},\label{eq:mom_flow_map}
\end{equation}
which is the flow map (starting from time $0$) corresponding to the Hamiltonian $H_{\mom, t}((q, p)) = \dot V_\mom(t,p)$. As above, the subscript m indicates that the Hamiltonian depends on momentum but not position. Although the Hamiltonians above take a very particular form, they naturally arise when applying splitting integration methods to separable Hamiltonians. By parametrising $V_\pos$ and $V_\mom$ as multi-layer perceptrons (MLPs), say, and composing such steps, we get a time-dependent symplectic map, the parameters of which can be optimised to fit data or, more generally, minimise an objective function. 
%\subsection{A backward error analysis for SympFlow}
\subsection{The Hamiltonian of the SympFlow architecture}\label{sec:hamSympFlow}
As we will see in Proposition \ref{pr:symplectic}, we can find a time-dependent Hamiltonian function corresponding to the SympFlow architecture. This allows us to essentially perform a \emph{backward error analysis}: while SympFlow does not generally solve the true ODE under consideration, it solves a time-dependent Hamiltonian ODE, the Hamiltonian of which we can give an expression for. We can apply the following result for this purpose:
\begin{proposition}[Proposition 1.4.D from \citet{polterovichGeometryGroupSymplectic2001}]
Let $H^1,H^2:\R\times \R^{2d}\to\R$ be continuously differentiable functions, and $\phi_{H_t^1,t},\phi_{H_t^2,t}:\R^{2d}\to\R^{2d}$ the exact flows (starting from time $0$) of the Hamiltonian systems they define. Then, the map $\psi_t=\phi_{H_t^2,t}\circ \phi_{H_t^1,t}:\R^{2d}\to\R^{2d}$ is the exact flow of the time-dependent Hamiltonian system defined by the Hamiltonian function
\[
H^3_t(x)=H^2_t(x)+H^1_t\Big(\phi_{H_t^2,t}^{-1}(x)\Big).
\]
\label{pr:symplectic}
\end{proposition}
\vspace{-2em}
As a result, given an overall SympFlow of the following form (with $\phi_{\mom, t}^i$ taking the form of \cref{eq:mom_flow_map} for some $V_\mom^i$ and $\phi_{\pos, t}^i$ taking the form of \cref{eq:pos_flow_map} for some $V_\pos^i$)
\begin{equation}\bar{\psi}_t = \phi_{\mom, t}^L \circ \phi_{\pos, t}^L\circ \ldots \circ \phi_{\mom, t}^1 \circ \phi_{\pos, t}^1,\label{eq:sympflow}
\end{equation}
we can associate the SympFlow with a time-dependent Hamiltonian. To denote this Hamiltonian, we introduce the operator $\mathcal{H}$ sending a SympFlow into one of its generating Hamiltonian functions (all of which differ just by a constant), so one has $\mathcal{H}(\bar{\psi}):\R\times\R^{2d}\to\R$.
%\begin{align*}
%H_{\mathrm{SympFlow},t} &=H_{\mom, t}^L(x) + H_{\ldots}([\phi_{\mom, t}^L]^{-1}(x))\\
%&= H_{\mom, t}^L(x) + H_{\pos, t}^L ([\phi_{\mom, t}^L]^{-1}(x)) + H_{\ldots}([\phi_{\mom, t}^L \circ \phi_{\pos, t}^L]^{-1}(x))
%\end{align*}
%{\color{red} Find a nice way to write it out}

To assemble such a function, we can group the pairs of alternated momentum and position flows, finding the Hamiltonian associated with $\phi_{\mom,t}^i\circ \phi_{\pos,t}^i$, which is $
H^i_t((q,p))=\dot{V}_{\mom}^i(t,p) + \dot{V}_{\pos}^i(t,q - (\nabla_p V_{\mom}^i(t, p) - \nabla_p V_{\mom}^i(0, p))).$ 
The Hamiltonian $\mathcal{H}(\bar{\psi})$ can then be expressed iteratively, aggregating from last layer to first as
\[
H_t^{L:i}(x)=H_t^{L:(i+1)}(x)+H_t^i\left(\phi_{H_t^{L:(i+1)}, t}^{-1}(x)\right)
,\,\,i=1,\ldots,L-1,
\]
where $H_t^{L:L} = H_t^L$, and
\[
\phi_{H_t^{L:i}, t}^{-1} = \left(\phi_{H_t^{L:(i+1)}, t}\circ \phi_{H_t^{i}, t}\right)^{-1} = \phi_{H_t^{i}, t}^{-1} \circ \phi_{H_t^{L:(i+1)}, t}^{-1}.
\]
The Hamiltonian of the network $\mathcal{H}(\bar{\psi})$ then corresponds to $H_t^{L:1}$. 
\vspace{-0.5em}
% The Hamiltonian $\mathcal{H}(\bar{\psi})$ can then be expressed iteratively, aggregating from left to right.  We define such a function up to the $i-$th pair of layers as
% \[
% \bar{H}_{t}^{i}(x)=\bar{H}_{t}^{i+1}(x) + H^i_t\Big(\phi_{\bar{H}_{t}^{i+1},t}^{-1}(x)\Big),\,\,i=1,\ldots,L-1,
% \]
% where $\bar{H}_t^L=H_t^L$. The Hamiltonian $\mathcal{H}(\bar{\psi})$ then corresponds to $\bar{H}^1_t$.

\subsection{Training SympFlow: Flow learning with Hamiltonian matching}
The natural task to use the SympFlow architecture for is to approximate the flow map of a Hamiltonian system as in \cref{eq:hamiltonian_ode}. We will assume the existence of a compact subset $\Omega\subset\R^{2d}$ which is forward invariant, meaning that $\phi_{H,t}(\Omega)\subseteq \Omega$ for every $t\geq 0$.

To train SympFlow, we consider a loss function composed of two terms. The first is the usual physics-informed (PI) loss function \citep{raissiPhysicsinformedNeuralNetworks2019}, based on the residual of \cref{eq:hamiltonian_flow_map},
\[
\mathcal{L}_1(\bar{\psi})=\frac{1}{N}\sum_{i=1}^N\Big\|\frac{\dd}{\dd t}\bar{\psi}_{t_i}(x_0^i) - \mathbf{J}\nabla H(\bar{\psi}_{t_i}(x_0^i))\Big\|_2^2,
\]
where $x_0^i\in \Omega$ and $t_i\in [0,\Delta t]$ for every $i=1,\ldots,N$. Unlike classical numerical integrators, $\Delta t $ can be chosen to be large. Using the analysis in the previous section, it is also possible to extract the underlying Hamiltonian of a given instance of SympFlow. This gives rise to a natural training objective, which we call the \emph{Hamiltonian matching loss} and which constitutes the second term in our loss, defined as
\[
\mathcal{L}_2(\bar{\psi})=\frac{1}{M}\sum_{i=1}^M \left(\mathcal{H}(\bar{\psi})(t_i,x^i) - H(x^i) \right)^2,
\]
where $x^i\in \Omega$ and $t_i\in [0,\Delta t]$ for every $i=1,\ldots,N$. The loss function we optimise over the space of SympFlows, is then $\mathcal{L}=\mathcal{L}_1+\mathcal{L}_2$, and we use PyTorch \citep{paszke_pytorch_2019} to get its gradient for use in the Adam optimiser \citep{kingma_adam_2017}.

Given a trained SympFlow network, representing a function $\bar{\psi}:[0,\Delta t]\times\R^{2d}\to\R^{2d}$, we can apply it to longer time horizons as follows: extend $\bar{\psi}$ to $[0,+\infty)$ via $\psi:[0,+\infty)\times \R^{2d}\to\R^{2d}$ defined as 
\begin{equation}\label{eq:longTimeExtension}
\psi_t(x_0):=\bar{\psi}_{t-\Delta t\lfloor t/\Delta t\rfloor}\circ \left(\bar{\psi}_{\Delta t}\right)^{\lfloor t/\Delta t\rfloor}(x_0),
\end{equation}
which can be considered an approximation of $\phi_{H,t}(x_0)$ for every $t\geq 0$.
\section{Experiments}
\label{sec:experiments}
We consider two Hamiltonian systems: the harmonic oscillator, with $H(q, p) = (q^2 + p^2)/2$, and the H\'enon-Heiles system, with $H(q_1, q_2, p_1, p_2) = (p_1 ^2 + p_2^2)/2 + (q_1^2 + q_2^2)/2 + q_1^2q_2 - q_2^3/3$. The H\'enon-Heiles system is notable for exhibiting chaos. We train SympFlow and a comparable MLP (using only physics-informed loss), on an interval with $\Delta t=1$. We also compare to an adaptive integrator, ODE45, as implemented in SciPy (with the default tolerances) \citep{2020SciPy-NMeth}. More results are shown in \cref{sec:appendix_experiments}, but note in particular the good long-time energy behaviour of SympFlow, compared to the other methods:
\begin{figure}[!htb]
\centering
\begin{minipage}{.5\textwidth}
  \centering
  \includegraphics[scale=1]{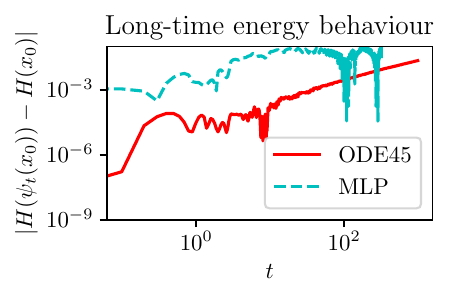}
\end{minipage}%
\begin{minipage}{.5\textwidth}
  \centering
  \includegraphics[scale=1]{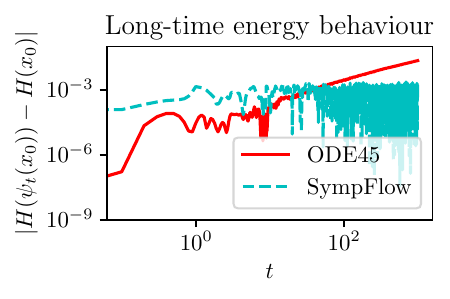}
\end{minipage}
\caption{A comparison of the long-time energy conservation of SympFlow and an unconstrained neural flow map approximator on the H\'enon-Heiles system.}
\label{fig:energy_hh}
\end{figure}
\vspace{-2.5em}
\section{Conclusions and discussion}
We have given a demonstration of the SympFlow architecture and its application to integrating Hamiltonian systems. As shown in our experiments in \cref{sec:experiments}, SympFlow exhibits good long-time conservation of energy, as is common for classical time-stepping symplectic numerical integrators as well. There is a large body of theoretical research on such properties for time-stepping integrators, which may be leveraged in future research to theoretically support the behaviour that we have observed for SympFlow. The setting of Hamiltonian systems is not as restrictive as it may appear at a first glance: we have only considered ODEs here, but extensions are possible to PDEs \citep{bridgesNumericalMethodsHamiltonian2006}, and even to non-conservative systems \citet{Galley_2013, Galley_2014, Tsang_2015}. Studying such extensions in more detail is a promising avenue for future research.

\bibliography{references}

\newpage
\appendix
\section{Additional experimental results}
\label{sec:appendix_experiments}
\subsection{Harmonic oscillator}
\begin{figure}[!htb]
\centering
\includegraphics[scale=1]{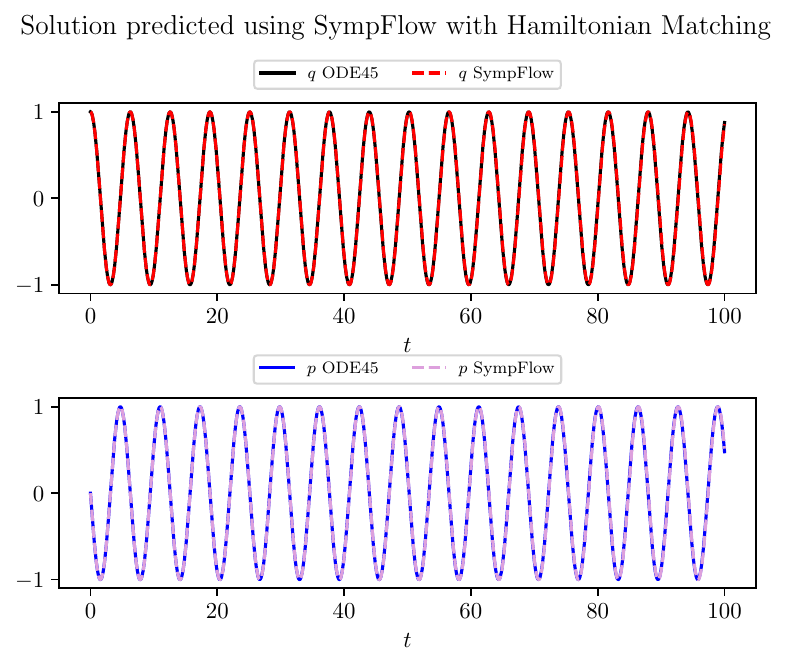}
\caption{A trajectory of the harmonic oscillator predicted by SympFlow and ODE45.}
\label{fig:traj_ho}
\end{figure}
\begin{figure}[!htb]
\centering
\begin{minipage}{.5\textwidth}
  \centering
  \includegraphics[scale=1]{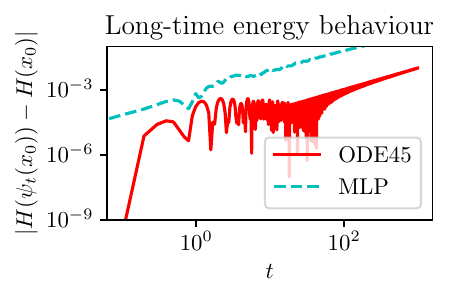}
\end{minipage}%
\begin{minipage}{.5\textwidth}
  \centering
  \includegraphics[scale=1]{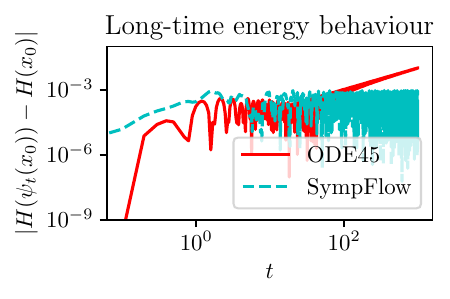}
\end{minipage}
\caption{A comparison of the long-time energy conservation of SympFlow and an unconstrained neural flow map approximator on the harmonic oscillator.}
\label{fig:energy_ho}
\end{figure}
\FloatBarrier
\subsection{H\'enon-Heiles system}
\begin{figure}[!htb]
\centering
\includegraphics[scale=1]{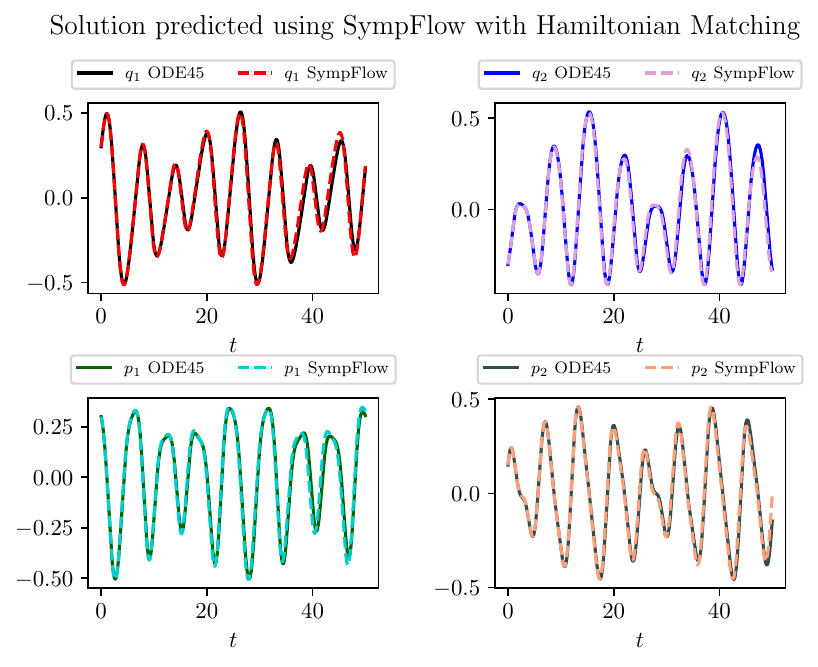}
\caption{A trajectory of the H\'enon-Heiles system predicted by SympFlow and ODE45.}
\label{fig:traj_hh}
\end{figure}
\end{document}